\documentclass[10pt,twocolumn,letterpaper]{article}

\usepackage[pagenumbers]{cvpr} % To force page numbers, e.g. for an arXiv version

\makeatletter
\@namedef{ver@everyshi.sty}{}
\makeatother

\usepackage{graphicx}
\usepackage{amsmath}
\usepackage{amssymb}
\usepackage{booktabs}

\usepackage{multirow}
\usepackage[table]{xcolor}
\usepackage{booktabs}
\usepackage{amsmath}
\usepackage{float}

\usepackage{cite}
\usepackage{amsmath,amssymb,amsfonts}
\usepackage{algorithmic}
\usepackage{graphicx}
\usepackage{textcomp}

\usepackage{multirow}
\usepackage{caption}
\usepackage{subcaption}
\usepackage{placeins}

\usepackage{listings}

\usepackage[pagebackref,breaklinks,colorlinks]{hyperref}

\usepackage[capitalize]{cleveref}
\crefname{section}{Sec.}{Secs.}
\Crefname{section}{Section}{Sections}
\Crefname{table}{Table}{Tables}
\crefname{table}{Tab.}{Tabs.}

\usepackage{svg}
\usepackage{amsmath}

\def\cvprPaperID{4698} % *** Enter the CVPR Paper ID here
\def\confName{CVPR}
\def\confYear{2022}

\begin{document}

\title{StereoSpike: Depth Learning with a Spiking Neural Network}

\author{Ulysse Rançon, \hspace{0.5cm} Javier Cuadrado-Anibarro, \hspace{0.5cm} Benoit R. Cottereau, \hspace{0.5cm} Timothée Masquelier\\
CerCo CNRS UMR 5549, Université Toulouse III\\
Toulouse, France\\
{\tt\small \{ulysse.rancon, javier.cuadrado, benoit.cottereau, timothee.masquelier\}@cnrs.fr}
}

\maketitle

%%%%%%%%% ABSTRACT
\begin{abstract}
   Depth estimation is an important computer vision task, useful in particular for navigation in autonomous vehicles, or for object manipulation in robotics. Here, we propose to solve it using \emph{StereoSpike}, an end-to-end neuromorphic approach, combining two event-based cameras and a Spiking Neural Network (SNN) with a modified U-Net-like encoder-decoder architecture. More specifically, we used the Multi Vehicle Stereo Event Camera Dataset (MVSEC). It provides a depth ground-truth, which was used to train StereoSpike in a supervised manner, using surrogate gradient descent. We propose a novel readout paradigm to obtain a dense analog prediction --the depth of each pixel-- from the spikes of the decoder. We demonstrate that this architecture generalizes very well, even better than its non-spiking counterparts, leading to near state-of-the-art test accuracy. To the best of our knowledge, it is the first time that such a large-scale regression problem is solved by a fully spiking neural network. Finally, we show that very low firing rates (\textless 5\%) can be obtained via regularization, with a minimal cost in accuracy. This means that StereoSpike could be efficiently implemented on neuromorphic chips, opening the door for low power and real time embedded systems.
\end{abstract}

%%%%%%%%% BODY TEXT
\section{Introduction}
\label{sec:intro}
Depth is an important feature of the surrounding space whose estimation finds its place in various tasks across many different fields \cite{stereo_biomedical}. Potential applications can be as diverse as object manipulation in robotics \cite{williams2019robotic} or collision avoidance for autonomous vehicles during navigation \cite{Neven2017FastSU}. In humans, depth processing is extremely well developed and relies on monocular (e.g., occlusions, perspectives or motion parallax) and binocular (retinal disparities) visual cues \cite{CuttingVishton}. This processing consumes very little energy as the visual system encodes retinal information under the form of action potentials, or \emph{spikes} and it is believed that the brain only requires about 20 Watts to function \cite{Mink1981}. Over the last years, it has motivated the development of numerous bio-inspired approaches based on neuromorphic sensors and spiking neural networks to process depth in embedded systems.

Dynamic Vision Sensors (DVS) have recently gathered the interest of scientists and industrial actors, thanks to a growing number of research papers explaining how to process their output \cite{Gallego2019}. Notable reasons for this recent popularity are their very high dynamic range and excellent temporal resolution which allow them to operate in extreme conditions (e.g., night, bright sun, rapid motion) where conventional frame-based cameras would suffer from severe saturation or motion blur.
Instead of repeating redundant frame information (i.e., when the camera or the scene is not moving) at a fixed sampling rate, their pixels asynchronously emit an action potential --\emph{spike} or \emph{event}-- whenever the change in log-luminance at this location since the last event, reaches a threshold. This sparse encoding scheme draws direct inspiration from retinal ganglion cells in animal models.
Finally, event cameras' unrivaled energy-efficiency also contribute to make them especially suitable in automotive scenarios with strong energy, memory and latency constraints.

Spiking Neural Networks (SNNs) are a good fit for DVSs, as they can leverage the sparsity of their output event streams. Implemented on dedicated chips such as Intel Loihi \cite{Loihi}, IBM TrueNorth \cite{Truenorth}, Brainchip Akida \cite{AkidaBrainchip} or Tianjic \cite{Pei2019}, these models could become a new paradigm for ultra-low power computation in the coming years. In addition, SNNs maintain the same level of biological plausibility as silicon retinae, making them new models of choice among computational neuroscientists. SNNs have recently attracted the deep learning community since the breakthrough of Surrogate Gradient (SG) learning \cite{SGlearning} \cite{Zenke2021}, which enabled the training of networks with back-propagation despite the non-differentiable condition for spike emission. While SNNs generally remain less accurate than their analog counterpart (i.e., Analog Neural Networks or ANNs), the gap in accuracy is decreasing, even on challenging problems like ImageNet \cite{fang2021deep}.
In this context, we bring the following contributions: 
\begin{itemize}
    \item We propose an ultra-low power spiking neural network for depth estimation, capable of dense depth predictions even at places without events and with high performances. In addition to its superior hardware-friendliness, our network is conceptually simpler than prior works and paves the way for using strictly spiking neural networks in other large-scale regression problems.
    \item We show that despite the dynamic nature of DVS data, the problem of depth estimation from such neuromorphic event streams can be treated as a non-temporal task. We leverage this feature by designing our model purely stateless in the sense that we reset all neurons to a membrane potential of zero at each timesteps. If this choice might not fully take advantage of the temporal processing abilities of SNNs, it drastically decreases the computational and energy footprint of our approach. 
    \item We report, to the best of our knowledge, one of the first SNNs outperforming equivalent ANNs in a serious and applied engineering task.
    \item We accompany this network with a new data augmentation technique for sequential DVS data which we call \emph{time mirror}.
\end{itemize}

In section \ref{related_work}, we introduce several related works that inspired our approach. We explain our methodology in section \ref{method}, including the data pre-processing, network architecture, and training details. In section \ref{experiments}, we compare our method with prior studies in terms on performances. Very interestingly, we also show that StereoSpike surpasses equivalent analog Neural Networks (ANNs) with similar architectures. Finally, section \ref{discussion} discusses the superiority of our model in terms of computational efficiency and energy-consumption.

\section{Related Work} \label{related_work}

Deep learning approaches for depth estimation have had a long tradition on the timescale of modern deep learning techniques. First methods were based on luminance-field data from traditional frame-based cameras, either in mono- or binocular setups. The model in \cite{Eigen} was the first successful multi-scale architecture designed for depth estimation from RGB images, and was consequently followed by advances based on similar approaches \cite{DispNet} \cite{Li2018MegaDepthLS} \cite{UnsupervisedLRconsistency}. 

Consistently with the recent interest of the scientific community in event-based cameras, a few works successfully tackled the problem with neuromorphic data. Historically, several groups used bio-inspired approaches such as Spiking Neural Networks (SNNs), in a very hardware-oriented direction, but not in a ``deep learning'' setting. For instance, the authors of \cite{indiveri} implemented a spike-based algorithm on a FPGA to regress low-resolution depth maps on a small size dataset. Furthermore, \cite{benosman} proposed a SNN for processing depth from defocus (DFD); this work targeted neuromorphic chips and was able of recovering depth at full resolution, but reconstructions were not dense and the approach was not based on learning.

Following the latter, \cite{DepthEgomotion} proposed a neural network that jointly predicted camera pose and per-pixel disparity from stereo inputs. However, their reconstructed depth maps remained sparse as they restricted their analyses to pixels where events occurred. \cite{tulyakov-et-al-2019} addressed this problem and pushed even further the state-of-the-art on indoor scenarios thanks to 3D convolutions exploiting a specific event embedding. 
The model in \cite{EITNet} used the same input embedding and backbone as \cite{tulyakov-et-al-2019}, but proposed a preliminary network using spatially-adaptive normalization (SPADE) \cite{park2019SPADE} to reconstruct grayscale intensity images jointly with depth maps. If the performance gain in this case is indeed important, this paradigm is very intensive as it adds much more parameters and FLOPs but also imposes to have access to ground-truth intensity images for training.
\cite{DTC_SPADE} also used a similar matching backbone as well as SPADE, but differed in its input encoding. In this work, subsequent spike histograms are fed sequentially into a layer of recurrent, non-spiking Leaky-Integrate and Fire (LIF) neurons with different time constants to capture time dependencies at different scales. Although this approach can be considered the current state-of-the-art of stereo matching DVS events, this model is difficult to implement on neuromorphic chips, as its activations are dense and not binary.
Finally, in \cite{Hidalgo20threedv}, dense metric depth was recovered from only one camera, and showed good performances with a recurrent, monocular encoder-decoder architecture on outdoor sequences. However, we argue that the task of depth recovery from events has a minor temporal component and can be solved by a fully feedforward model with minimal temporal knowledge; therefore the use of convLSTMs in \cite{Hidalgo20threedv} is suboptimal and unnecessarily costly in terms of computation. 

Another inspiration for our work has been the task of optical flow regression from neuromorphic data, which is similar to depth reconstruction because it is also a large-scale image regression task. 
Despite this similarity, it inspired lighter and hardware-friendly approaches, closer to the philosophy of SNNs, but still no fully spiking --to the strict sense-- models have been proposed for this purpose.
EV-FlowNet \cite{EVflownet}, arguably considered as the precursor of encoder-decoder models for optical flow reconstruction from event data, consisted in a feedforward analog encoder-decoder architecture. As a direct sequel, the hybrid model Spike-FlowNet \cite{spikeflownet} used spiking neurons in the encoder of a similar backbone, while maintaining the same levels of performances. In this approach, spiking neurons were shown to be able of encoding abilities close to analog ones and with a reduced computational cost. On the other hand, authors kept the remaining part of their network analog, to counteract the lack of expressivity in SNNs. More recently, the model proposed in \cite{paredesvalles2021selfsupervised} showed very good performances but it cannot be considered as a fully spiking network because real-valued intermediate predictions of the outputs were reinjected within the network and mixed with binary spike tensors. In addition, they upsampled low-scale representations with the bilinear upsampling method, which breaks the binary spike constraint necessary for an implementation on neuromorphic hardware. Nevertheless, it is the first success in a large-scale regression task with a network that is spiking for its vast majority.

So far, SNNs have been used for classification tasks like image recognition \cite{Fang2021,fang2021deep}, object detection \cite{Kim_Park_Na_Yoon_2020,Cordone2022}, or motion segmentation \cite{spikeMS}. Only a few works employed them for regression tasks. A notable exception is \cite{angularSNN}, but they only regressed 3 variables, while we propose here to regress the values of $260\times346=89960$ pixels.

\section{Method} \label{method}

We used PyTorch and SpikingJelly \cite{SpikingJelly} as our main development libraries. PyTorch is currently one of the most popular tools for deep learning and automatic differentiation, while Spikingjelly is an open-source framework for spiking neural networks, based on PyTorch and with rising popularity. 
Our codes are partially available on GitHub at the following address: \url{https://github.com/urancon/stereospike}. We plan for a full release upon publication of the paper. Within this repository is a link to a Weights and Biases \footnote{\url{https://wandb.ai/}} report, compiling trainings that led to the set of hyperparameters we present in this paper.

\subsection{Dataset} \label{dataset}

We trained and tested our network on the Multi Vehicle Stereo Event Camera (MVSEC) dataset \cite{mvsec}. Because of its large size and variability, it has become one of the most popular benchmarks for depth reconstruction from neuromorphic events. It was collected from two DAVIS346 cameras with a resolution of $346\times260$ pixels, mounted on several vehicles such as a car, a motorbike or a drone. The depth groundtruth was provided by a Velodyne Puck Lite LIDAR mounted on the top of the two event cameras and with a sampling frequency of 20 Hz, hence providing a ground truth depth map every 50 ms. 

We first applied our method on the \textit{indoor\_flying} sequences of MVSEC, which was recorded on a quadricopter flying inside a large room. We used the data splits that were defined in \cite{TSES} and \cite{tulyakov-et-al-2019}. We followed these previous works and removed take-off and landing parts of the sequence, because they contained very noisy event streams and inaccurate groundtruths. As a result, indoor settings were represented on average by 2850, 200 and 1100 samples for training, validation and test sets respectively.

We then investigated outdoor scenarios by following the same training, test and validation splits as \cite{Hidalgo20threedv}. Namely, the largest \textit{outoor\_day2} sequence was used for training and validation, and testing was performed on one day-time and one night-time sequences, respectively \textit{outoor\_day1} and \textit{outoor\_night1}.
Please note that \cite{Hidalgo20threedv} also used simulated data that did not come from MVSEC, but this does not prevent StereoSpike from outperforming this competitor by a large margin with substantially less parameters (cf. Table \ref{table:perfs}). In summary, we used respectively 8520, 1820 and 5130 samples for training, validation and testing outdoor conditions.

\subsection{Event Representation} \label{event representation}

\begin{figure}
    \centering
    \includegraphics[width=0.45\textwidth]{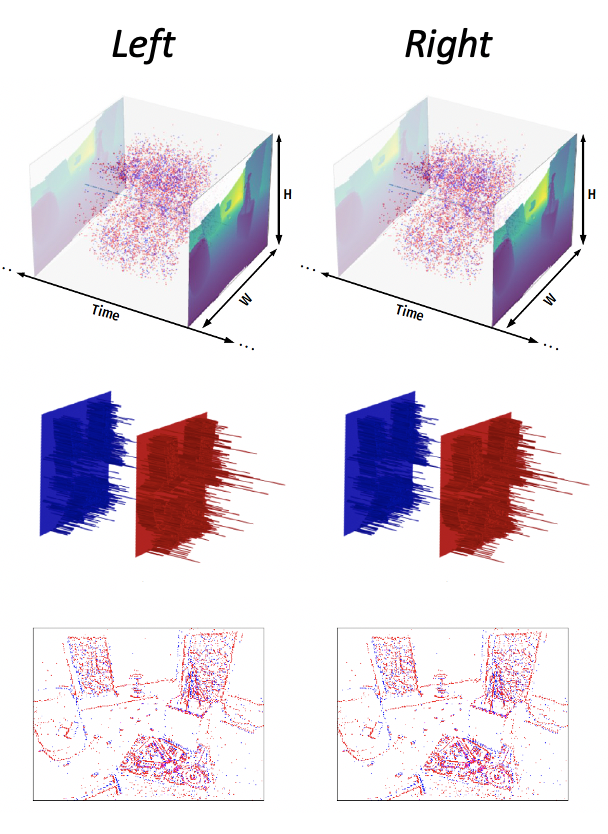}
    \caption{(top): ON and OFF events are binned, per-pixel, within time windows of 50 ms. Frames displayed on the temporal axis are the ground-truth depth maps, provided by the LIDAR at 20 Hz. (middle): This operation results in a 2-channel histogram of spikes, containing integer spike counts at every pixel and for each polarity. (bottom): Such spike frames are commonly visualized according to the following convention: pixels reporting at least one ON event are colored in red, those reporting OFF events in blue, and those reporting both types in pink.}
    \label{fig:fig1}
\end{figure}

We adopted a rather common representation of data: we binned all incoming spikes on each pixel on a time window of 50 ms (see Figure \ref{fig:fig1}). Furthermore, we accumulated spikes for each polarity in a different channel. Because there are two polarities, the resulting tensor had a shape of $(2, Height, Width)$ and contained positive integers, corresponding to the number of spikes of each polarity that showed up at each position of the scene during the time window. We further refer to this format as \emph{spike histograms} or \emph{spike frames} interchangeably.

The duration of 50 ms for binning was motivated by empirical results, as this value leads to better model performances than durations of 25 ms or 100 ms.  

The final input tensor is obtained by concatenating the spike frames from left and right cameras together channel-wise, hence resulting in a $(4, Height, Width)$-shaped volume.

Many ANN approaches normalize this kind of input tensor (e.g. divide by the maximum number of spike count) for an easier-to-learn distribution of data and better generalization \cite{Hidalgo20threedv}. We believe that this operation has a high cost on neuromorphic hardware, and can lead to non-integer number of spikes in the normalized input tensor. For this reason, we prefer feeding raw spike frames directly to our network. As a counterbalance measure, we used data augmentation; as \cite{Hidalgo20threedv}. Pretraining the model on simulated data may also help generalize to other distributions.

\subsection{Neuron model}

We use the McCulloch and Pitts model \cite{mcculloch1943logical}, outputting a binary activation when the amount of weighted spikes integrated from lower layers reaches a threshold:

\begin{equation}
\label{eq1}
    S^{l} = \Theta(V_{reset}^{l} + \sum_{}^{}w^{l-1}*S^{l-1})
\end{equation}

where $\Theta$ is the Heaviside step function, $l$ denotes the layer number, and $w$ synapse weights. $V_{reset}$ corresponds to the potential of neurons at rest, and acts as an offset -- a bias-- to facilitate or hinder neurons from spiking.

This model is equivalent to the Integrate-and-Fire (IF) model deprived of the implicit recurrence in the membrane potential; all neuron potentials are reset to a value of $V_{reset}$ at every timestep. As a result this model is stateless, contrarily to the traditional IF model. We use SpikingJelly's \textit{IFNode} class for its implementation.
Such neurons are inexpensive to simulate and can be deployed in large models, whereas more complex ones such as Hodgkin-Huxley \cite{Hodgkin1952}, Izhikevich \cite{Izhikevich2003} or even SRM \cite{Gerstner1993} are still too computationally expensive to be trained on modern hardware.

A problem that has long prevented researchers from using simple Integrate-and-Fire models with standard deep learning techniques, is the gradient of the activation function --the Heaviside function-- being zero everywhere (except in 0 where it is not defined). A recent solution to this is the replacement of the true gradient by a surrogate \cite{SGlearning}, which lets more room for the gradient to flow. We use the derivative of the $\arctan$ function as our surrogate gradient in this paper, as suggested in \cite{Fang2021}.

\subsection{Architecture}

\begin{figure*}[ht]
    \centering
    \includegraphics[width=\textwidth]{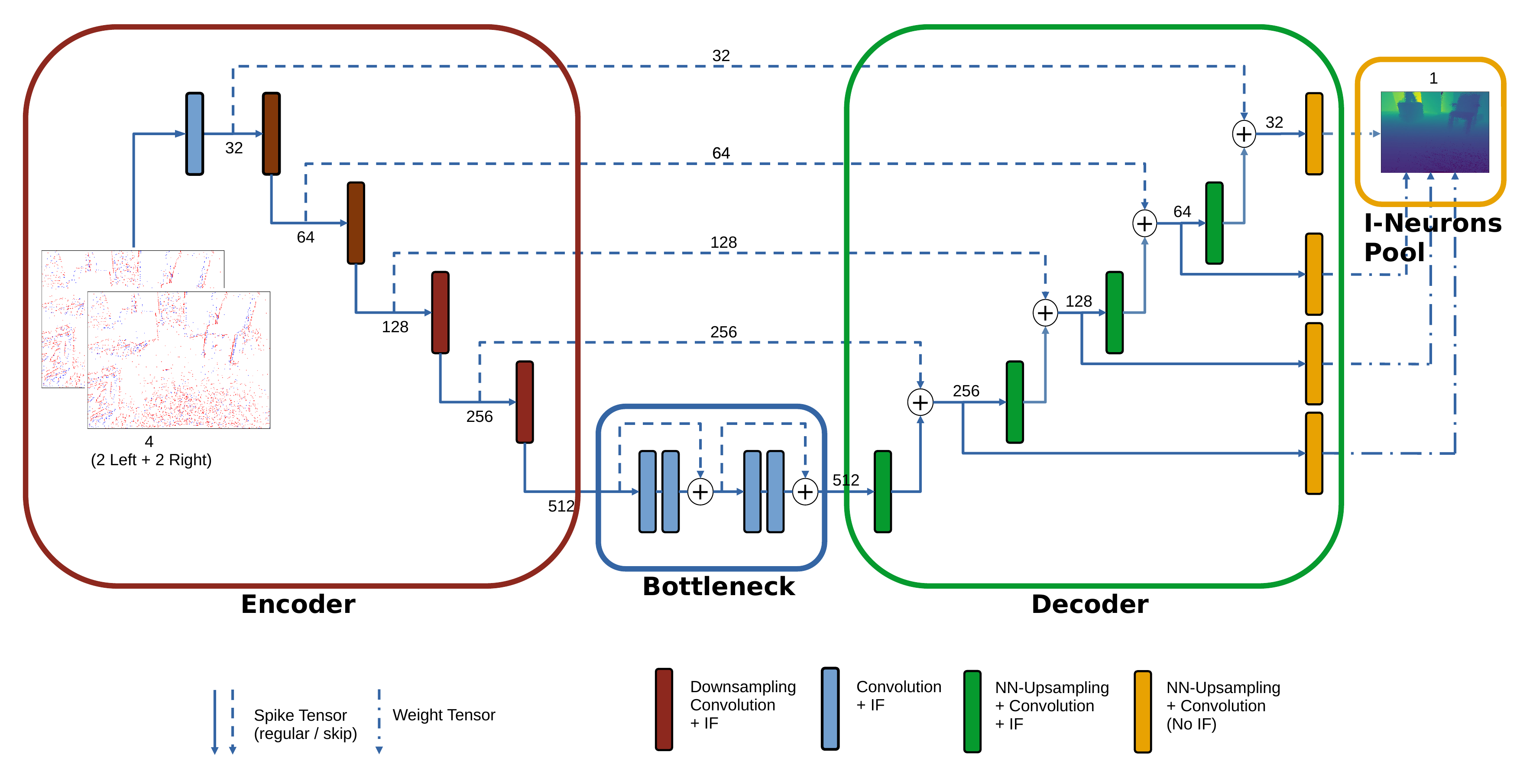}
    \caption{Detailed architecture of StereoSpike. Its first convolutional layer has 2 or 4 input channels (depending on the modality), essentially one pair for a each event camera. These encoded spike frames are further processed through a bottleneck consisting of 2 chained SEW-Resblocks. As a result, the tensor out of these residual layers is composed of integers in range [0, 3]. This latent representation is progressively upsampled by decoder layers and whose outputs are summed with same-level encoder spike tensors, leading to integer values in range [0, 2]. In parallel, prediction synapses from different scales directly project to output I-neurons whose membrane potentials bear the final prediction. The numbers indicate the size of channel dimension for each spike volume. Best viewed in color.}
    \label{fig:archi}
\end{figure*}

Our model is fully convolutional and based on a U-Net backbone \cite{DBLP:journals/corr/RonnebergerFB15} consisting of an encoder, a bottleneck and a decoder whose non-linearities were achieved by spiking neurons (see Figure \ref{fig:archi}). We used the McCulloch and Pitts model presented in (\ref{eq1}) with a reset potential of $V_{reset}=0$. \\

Downsampling in the encoder was performed by 2-strided convolutions, which divided the spatial resolution by 2 while doubling the channel resolution. The bottleneck consisted in 2 SEWResBlocks \cite{fang2021deep} following each other and with $ADD$ connect function. 

Because transposed convolutions are known to generate checkerboard artifacts \cite{odena2016deconvolution}, decoder upsampling layers rather consist in nearest neighbor (NN) upsampling followed by a convolution. Contrarily to bilinear upsampling, we believe NN to be neuromorphic-hardware friendly, because it essentially keeps integer spike counts in the upsampled volume. In terms of biological plausibility, it can be viewed as a single neuron at low scale projecting synapses towards several higher-scale neurons.

The output of the network was carried by the potentials of a pool of non-leaky neurons with an infinite threshold. One critical problem of SNNs is their lack of expressivity, since they can only update the membrane potentials of output neurons with discrete weighted spikes. With synapses coming from the full scale level of the encoder only as in a standard architecture, the model would have too few spikes and too few different parameters (i.e., synaptic weights) to achieve top performances at rendering a large-scale and diverse depth scene.
To counteract this effect, we increased the number of spikes to update the readout neurons' potential by linking them to lower levels of the network \textit{via} intermediary prediction layers. These layers essentially consisted in nearest neighbor upsampling followed by convolution; they were the same as the upsampling layers in the decoder, except that they upsampled spike tensors directly up to the full, original scale.

In order to capture long-range spatial dependencies while maintaining a low amount of parameters, all convolutions in the model used large $7 \times 7$ kernels and were separable (depthwise followed by pointwise).
In addition, none of the layers in our network used a bias term nor Batch Normalization (BatchNorm) because adding constant biases is costly on neuromorphic hardware and is not biologically plausible.

\subsection{Loss Function}

As in \cite{Hidalgo20threedv}, we used a combination of a regression loss with a regularization loss. Noting $R=\hat{D}-D$ the residual between the groundtruth and predicted depth maps, the first term can be written as:

\begin{equation}
        L_{regression}=\frac{1}{n}(\sum_{u}^{}(R(u))^{2} - \frac{1}{n^{2}}(\sum_{u}^{}R(u))^{2})
\end{equation}

where $n$ is the number of valid groundtruth pixels $u$. With the same notations, the regularization loss is computed with:
\begin{equation}
    L_{smooth}=\frac{1}{n}\sum_{u}^{}|\nabla_{x}R_{s}(u)|+|\nabla_{y}R_{s}(u)|
\end{equation}

According to \cite{Hidalgo20threedv}, the minimization of this term encourages smooth depth changes as well as sharp depth discontinuities in the depth map prediction, hence helping the network to represent objects that stand out of the background of the scene (e.g., because they are closer), while respecting its overall topology. Finally, we weighted both terms with a factor $\lambda$ in the total loss:
\begin{equation}
    L_{tot}=L_{regression}+\lambda L_{smooth}
\end{equation}

We used a value of $0.5$ for $\lambda$, which was determined empirically.
This loss was applied on all intermediate predictions, therefore giving out 4 loss terms: the first being from the lowest prediction layer, and the last being the actual prediction and the sum of all 4 prediction layers. This encourages the network to predict relevant depth images as early as possible; we determined that this strategy on the application of the loss gives better results than simply applying it on the final prediction.

\subsection{Training Procedure} \label{training_procedure}

Parameter values in our model were learnt using the method of the surrogate gradient \cite{SGlearning}, as implemented in Spikingjelly Python library \cite{SpikingJelly}. 

Because our network is feed-forward and only processes one step for inference, we trained it on the shuffled dataset with regular back-propagation, not Back-Propagation Through Time (BPTT).

\begin{figure}[H]
    \centering
    \includegraphics[width=0.45\textwidth]{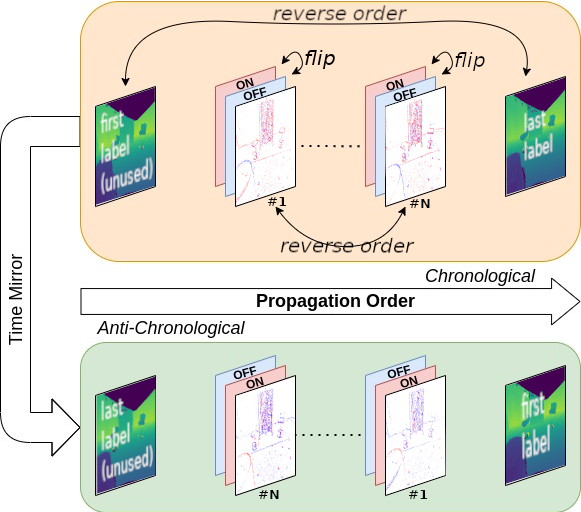}
    \caption{"Time Mirror" data augmentation technique. The events generated between the timestamps of two consecutive labels are cumulated into $N$ ON-OFF frames of temporal duration $T$ and concatenated channel-wise. In the case of this study, $N=1$ and $T=50ms$. A sequence in anti-chronological order is a piece of data as valid as the original in chronological order. To reverse the order of a DVS sequence, event timestamps must be reversed --therefore translating into an inversion of the order of the histograms-- but event polarities must be switched as well.}
    \label{fig:time_mirror}
\end{figure}

% ---- DATA AUGMENTATION PARAGRAPH ------
To avoid overfitting, we used random horizontal flip and another technique that we call \emph{time mirror}. Conceptually simple, it applies to sequential DVS data and consists in feeding the frames and label in anti-chronological order. Because event polarities indirectly carry temporal information, it is needed to switch both ON and OFF channels in spike frames. To our knowledge, such a technique is new. We present it in Figure \ref{fig:time_mirror} for the general case where the duration between two ground-truth is cut into $N>1$ histograms.

We used Adam optimizer \cite{Kingma2015AdamAM} with $\beta_{1}=0.9$ and $\beta_{2}=0.999$. We trained the network for 30 epochs with an initial learning rate set to $2.10^{-4}$ and divided by 10 at epochs 10, 25 and 40. From preliminary tests, we concluded that the optimal batch size was 1. Similarly,
we found that weight decays did not improve performances and thus did not use it.
The whole training process took  7 hours on average on a single Nvidia Titan V with 12 GiB VRAM capacity, with an actual memory consumption of around 2.5 GiB.

\section{Experiments} \label{experiments}

\subsection{Performances}

\begin{figure*}[ht]
    \centering
    \includegraphics[width=\textwidth]{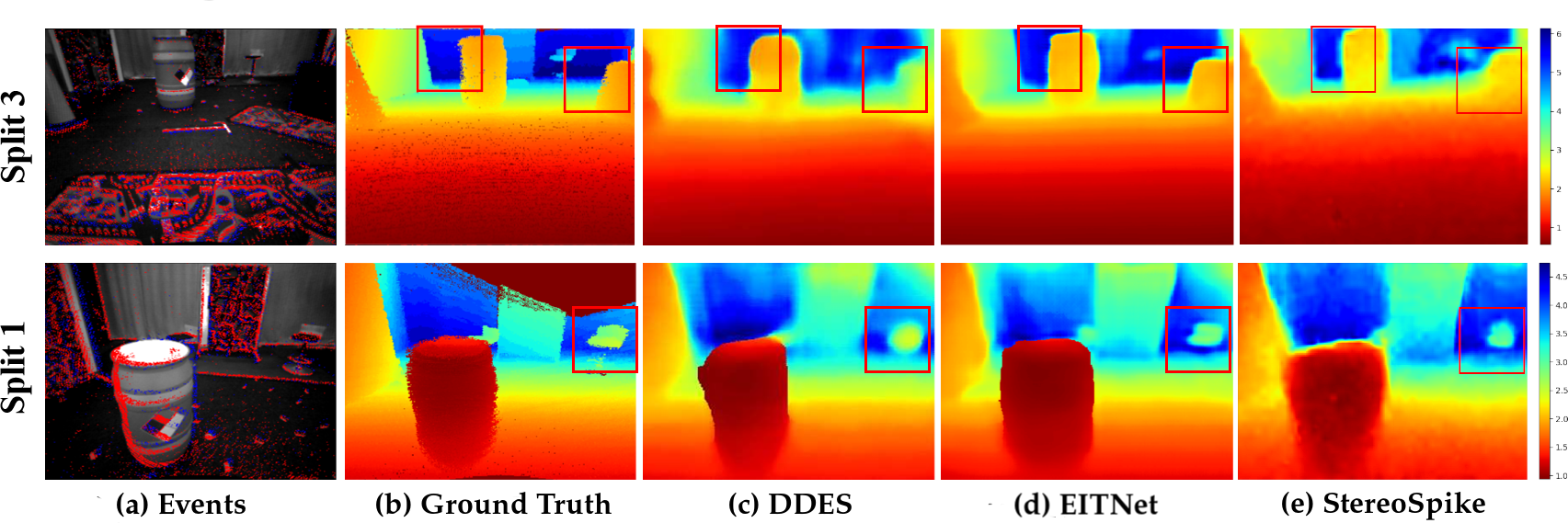}
    \captionsetup{justification=centering}
    \caption{Qualitative comparison of our method with other state-of-the-art approaches. We selected the same input frames and run our model to infer depth from this data; Event ground-truth and other prediction images were borrowed and adapted from\cite{tulyakov-et-al-2019} \cite{EITNet}. The top row corresponds to frame \#1700 of \textit{indoor\_flying3} and frame \#980 of \textit{indoor\_flying1} sequences. The pixelated aspect of our predictions comes from the Nearest-Neighbor interpolation in the prediction layer from very-low to full scale. Even so, it can be seen that our fully spiking network captures the scene just as well as a cutting edge ANN using a heavy framework and 3D convolutions. DTC-SPADE \cite{DTC_SPADE} is not represented here as the authors did not use the same colormap as other studies and did not publish their code. We tried but were still unable to faithfully reproduce their rendering. The above figure was obtained by using reversed JET colormap in the natural metric depth scale.} 
    \label{fig:predictions}
\end{figure*}

Figure \ref{fig:predictions} shows qualitative visualizations of depth reconstructions obtained with our model. Table \ref{table:perfs} provides a quantitative comparison with previous works on the Mean Depth Error (MDE), the most common metric used for characterizing depth estimation on MVSEC.

In outdoor scenarios, our model outperforms E2Depth \cite{Hidalgo20threedv} by a large margin at all cutoff distances, and with 25$\times$ fewer parameters. This is not due to the fact that \cite{Hidalgo20threedv} takes input from only one camera as the monocular version of StereoSpike still remains consistently superior to this competitor. 

In indoor settings, our model also proves to be better in terms of both  accuracy and number of parameters than the previous state-of-the-art (DDES, \cite{tulyakov-et-al-2019}), which is a fully-fledged ANN using 3D convolutions with a less general framework. 
However, recent approaches \cite{EITNet}\cite{DTC_SPADE} have built upon the latter and positioned themselves with lower MDE than StereoSpike. In the case of EITNet \cite{EITNet}, this improvement comes at the cost of a much higher number of parameters (more than 10 times). DTC-SPADE \cite{DTC_SPADE} has managed to incorporate elements of the latter for the benefit of accuracy and without exploding algorithmic size. Nevertheless, we argue that this model is heavier than StereoSpike and less suited to edge and energy-efficient computing in embedded systems.
With StereoSpike, we aim to address the problem of dense depth estimation from events consistently with the philosophy of event cameras, that is, by placing ourselves on the portability side of the accuracy-lightweightness trade-off. Moreover, unlike the recent state-of-the-art for this task \cite{tulyakov-et-al-2019} \cite{EITNet} \cite{DTC_SPADE}, StereoSpike's architecture is not specific to the estimation of depth and could be used for any other large-scale regression task; its simple adaptation from the standard 2D U-Net base makes it all the more general.

In terms of MDE, monocular models (i.e., receiving data from only one camera) also lead to good depth estimates, but with a consistent drop in accuracy across data splits.
This suggests that in addition to being --for the most part-- a non-temporal task, depth reconstruction from DVS data can be efficiently tackled on a monocular setting, at a reasonable cost in performance. Therefore, fusing left and right data as early as in the first convolution layer reveals itself to be a simple but good strategy for exploiting binocular disparity, as backpropagation proves capable of extracting stereoscopic cues in such a way. From a visual neuroscience perspective, there is strong evidence that binocular cues are mixed in a similar manner at very early stages of the visual  system\cite{Wandell1995FoundationsOV}. 

\begin{table*}[ht]
    \centering

    \begin{subtable}[h]{\textwidth}
        \centering
        \begin{tabular}{ccccccccccc} 
        \toprule
        \multirow{4}{*}{\textbf{Model}}             & \multicolumn{10}{c}{\textbf{ Mean Depth Error (MDE) [cm]}}                                                                                                                                                                            \\
                                                    & \multicolumn{2}{c}{\textbf{indoor}}         & \multicolumn{8}{c}{\textbf{outdoor}}                                                                                                                                                   \\
                                                    & \textbf{split 1}     & \textbf{split 3}     & \multicolumn{4}{c}{\textbf{outdoor\_day1}}                                                & \multicolumn{4}{c}{\textbf{outdoor\_night1}}                                               \\
                                                    & inf                  & inf                  & 10                   & 20                   & 30                   & inf                  & 10                   & 20                   & 30                   & inf                   \\ 
        \midrule
        DDES \cite{tulyakov-et-al-2019}                                        & 16.7                 & 27.8                 & -                    & -                    & -                    & -                    & -                    & -                    & -                    & -                     \\
        EITNet \cite{EITNet}                                      & 14.2                 & 19.4                 & -                    & -                    & -                    & -                    & -                    & -                    & -                    & -                     \\
        DTC-SPADE \cite{DTC_SPADE}                                  & \textbf{13.5}        & \textbf{17.1}        & -                    & -                    & -                    & -                    & -                    & -                    & -                    & -                     \\
        E2Depth \cite{Hidalgo20threedv}                                    & -                    & -                    & 1.85                 & 2.64                 & 3.13                 & -                    & 3.31                 & 3.73                 & 4.32                 & -                     \\
\rowcolor{lightgray}        StereoSpike - Binocular                     & 16.5                 & 18.4                 & \textbf{0.79}        & \textbf{1.47}        & \textbf{1.92}        & \textbf{3.17}        & \textbf{1.38}        & \textbf{2.26}        & \textbf{2.97}        & 4.82                  \\
\rowcolor{lightgray}        StereoSpike - Monocular                     & 18.6                & 28.6                 & 1.35                 & 2.30                 & 2.75                 & 4.01                 & 1.68                 & 2.61                 & 3.18                 & \textbf{4.46}         \\
        \bottomrule
        \end{tabular}
        \captionsetup{justification=centering}
        \caption{\textbf{Accuracy.} Test Mean Depth Error (MDE) in centimeters on several subsets of MVSEC. From three randomized training trials, the best model on the validation set is selected and then evaluated on the test set, from which these metrics are calculated. The number below the test set indicates the cut-off distance over which pixels are not taken into account for the MDE calculation. * indicates that the evaluation is done on the sparse groundtruth, i.e., only at pixels where events occurred.}
        \label{table:perfs}
    \end{subtable}
    
    ~\\~\\
    
    \begin{subtable}[ht]{\textwidth} % DEPLOYABILITY
        \centering      
        \begin{tabular}{cccccc}
        \toprule
        \textbf{model} & \textbf{GPU-compatible} & \textbf{Neuromorphic-compatible} & \textbf{Sparsity} & \textbf{Activation Width}  & \textbf{\# Params [M]}  \\
        \midrule 
        DDES \cite{tulyakov-et-al-2019}          & \textbf{Yes}            & No                               & No                & 32                         & 2.33                    \\
        EITNet \cite{EITNet}        & \textbf{Yes}            & No                               & No                & 32                         & 22.32                   \\
        DTC-SPADE \cite{DTC_SPADE}     & \textbf{Yes}            & No                               & No                & 32                         & 2.20                    \\
        E2Depth \cite{Hidalgo20threedv}       & \textbf{Yes}            & No                               & No                & 32                         & 43.22                   \\
\rowcolor{lightgray}        StereoSpike    & \textbf{Yes}            & \textbf{Yes}                     & \textbf{Yes}      & \textbf{1-2}               & \textbf{1.59}            \\
        \bottomrule
        \end{tabular}
        \captionsetup{justification=centering}
        \caption{\textbf{Deployability.} Set of features revealing the model computational efficiency and possible hardware implementation. In addition to being compatible with neuromorphic hardware, our model has sparse, low bit-width activations and requires significantly less parameters to achieve near state-of-the-art performances.}
        \label{table:deploy}
    \end{subtable}
    
    ~\\
    
    \captionsetup{justification=centering}
    \caption{Comparison of StereoSpike's performances with the state-of-the-art. Our philosophy values solving the accuracy-deployability trade-off, instead of the sole accuracy.}
\end{table*}

\subsection{Ablation studies}

\subsubsection{Intermediary predictions}

Contrarily to a standard U-Net \cite{DBLP:journals/corr/RonnebergerFB15}, StereoSpike makes a coarse-to-fine reconstruction of the depth scene \textit{via} four prediction layers, which can be seen as synapses projecting from different levels of the network body (i.e., decoder) to the pool of readout neurons. This technique is new, and to determine its added value, we conducted an ablation study on the architecture by ``cutting'' the latter and observing the performances of models partially deprived of their means of expressivity. Results are compiled in Table \ref{table:ablation}.\\

\begin{table}[H]
\centering
\begin{tabular}{l c}
\toprule
\textbf{Model}                      &        \textbf{MDE [cm]}     \\
prediction layer number             &                              \\
\midrule
\{1, 2, 3, 4\}                      &   \textbf{16.5   $\pm$ 0.3}                 \\
\{1, 2, 3\}                         &   18.1           $\pm$ 0.3                  \\
\{1, 2\}                            &   19.3           $\pm$ 0.5                  \\
\{1\}                               &   20.1           $\pm$ 0.8                  \\
\bottomrule
\end{tabular}
\caption{Test Mean Depth Error (MDE) on split 1 of \textit{indoor\_flying} sequence. Reported errors are averaged and provided with standard deviations over three randomized training trials. Prediction layers are depicted by a number $\in \{1, 2, 3, 4\}$, where 1 and 4 are the top- and bottom- level prediction layers, respectively.}
\label{table:ablation}
\end{table}

We observe that performances gradually decay as we remove intermediary prediction layers. Best performances are obtained by the full model with all four layers as presented in Figure \ref{fig:archi}, while the model equipped with top-level prediction layer only (i.e., classical encoder-decoder configuration) reports the worst metrics. Thus, multiplying regression layers constitutes an efficient strategy to enable more precise predictions.
More synapses means more parameters but also more possible spikes to update the potentials of output neurons, and therefore more degrees of freedom. This strategy is translated into an improved accuracy compared to standard encoder-decoder architectures, and could even be to analog models.

\subsubsection{Skip connections} \label{skip_co}

Skip connections are a standard feature of encoder-decoder and residual neural network architectures in the field of computer vision. However, they can be difficult to implement on some neuromorphic hardware. With the concern of staying close to low-level designers, we studied the effect of entirely removing them from StereoSpike (cf. Table \ref{table:ablation_skip}).

\begin{table}[H]
\centering
\begin{tabular}{lcc}
\toprule
\textbf{Model}                        & \multicolumn{2}{c}{\textbf{MDE [cm]}} \\
 Skip connections                     & train                       & test                          \\
\midrule
With                                  & 8.7  $\pm$ 0.1                 & \textbf{16.5} $\pm$ 0.3       \\
Without                               & \textbf{6.9} $\pm$ 0.7        & 17.1 $\pm$ 0.5                \\
\bottomrule
\end{tabular}
\caption{Test MDE on split 1 of \textit{indoor\_flying} sequence. Entries are averaged over three randomized training trials.}
\label{table:ablation_skip}
\end{table}

Models trained without skip connections turned out to overfit more than standard StereoSpike on training data, to the detriment of test accuracy. Therefore, skip connections seem to act like a regularizer here. However, the test accuracy drop is low, and a model without skip connections could still be reliably used in real situations. In the future, we believe that neuromorphic chips should be able to implement this typical architectural feature.

\subsection{StereoSpike - ANN comparison}

% also cite Federico, and maybe fusion-flownet
A lesson learnt from our study and \cite{spikeflownet} is that SNNs can encode information very optimally, even with binary values. While Spike-FlowNet used ANNs to decode the latent space representation, we only use spiking neurons. In addition, we do not mix real-valued intermediary predictions with integer spikes as in \cite{paredesvalles2021selfsupervised}. SNNs can therefore efficiently encode information as well as decode it, even for large scale regression tasks.

\begin{table}[H]
\centering
\begin{tabular}{lcccc}
\toprule
\multirow{2}{*}{\textbf{Model}}             & \multicolumn{2}{l}{\textbf{MDE [cm]}} & \multicolumn{2}{l}{\textbf{Loss [au]}} \\
                                             & train           & test           & train            & test           \\
\midrule
StereoSpike                                  & 8.7            & \textbf{16.5}  & 0.81             & \textbf{1.15}  \\
ANN (Sigmoid + BN)                           & 7.6            & 28.1           & 0.69             & 1.51           \\
ANN (Tanh + BN)                              & \textbf{6.9}   & 26.3           & \textbf{0.67}    & 1.50           \\
ANN (LeakyReLU + BN)                         & 8.2            & 26.0           & 0.68             & 1.42           \\
\bottomrule
\end{tabular}
\caption{Comparative evaluation of our SNN vs equivalent ANN models on split 1 of \textit{indoor\_flying} sequence. Entries are averaged over three randomized training trials. Our fully spiking network surpasses by a large margin all of its analog relatives. }
\label{table:snn_vs_ann}
\end{table}

In an attempt to compare our model with fully-fledged ANNs, we trained equivalent ANN models. These models had a exactly the same architecture and output paradigm consisting in a pool of I-neurons; however, and with the idea of using the full power of analog models, we replaced IFNodes by common activation functions, Nearest-Neighbor upsampling by bilinear upsampling, and used Batch Normalization (BN) \cite{Batchnorm} and trainable biases in convolution layers. As can be seen in Table \ref{table:snn_vs_ann}, the ANNs outperform the SNN on the training set, but not on the testing set. Specifically, the StereoSpike network --equivalent to an ANN with Heaviside step function as activation-- achieves the best test loss and MDE but also the worst training metrics, therefore generalizing best, despite the absence of BN. In other words, the ANNs overfit more than the SNN. This suggests that spikes, in addition to increasing hardware-friendliness, constitute an efficient regularization mechanism, causing the SNN to generalize better. To the best of our knowledge, it is the first time that this desirable regularization effect is reported - but of course, other regularization methods could be tried to limit overfitting in the ANNs (e.g. Dropout).

\subsection{Encouraging sparsity through spike penalization}
A promising feature of SNNs is their ability to solve complex tasks at performances comparable to conventional networks (ANNs), but with sparse activations. In order to quantify the computational efficiency of our network, we measure the firing rates of its layers, i.e., the density of intermediary tensors computed during inference on the test set. Sparse volumes can be leveraged by dedicated hardware capable of sparse computation, hence diminishing inference time as well as energy consumption. 

It appears that the firing rates of our best model grow as layers become closer to the output I-neuron pool. That is, convolution layers in the decoder report an average firing rate of 23.5\% compared to 8.1\% in the encoder. We suggest that in this large-scale regression task, a minimum number of pre-synaptic spikes preceding output neurons is necessary to faithfully render the visual scene. Similarly, a certain amount of spikes could be necessary to encode the information contained in the input histograms. 

To encompass this trade-off and estimate these minimal firing rates, we apply a regularization loss term explained in \cite{Pellegrini2021} and train a new binocular model on split 1. 
This secondary loss, which we also call \emph{quadratic spike penalization loss}, penalizes the mean of the squared spike tensor. Therefore, for a given layer containing K spiking units whose output at time-step $n$ is $S_{k}[n]\in \{0\ldots4\}$, it can be defined as:
\begin{equation}
\begin{split}
    L_{spikes} & = \frac{1}{2NK}\sum_{n}^{}\sum_{k}^{}S_{k}[n]^{2} \\
               & = \frac{1}{2K}\sum_{k}^{}S_{k}[n]^{2}
\end{split}
\end{equation}

Because the number of time-steps to do one prediction is $N=1$, as our model is purely stateless/feedforward. We apply this loss on the tensor out of the bottleneck and on the resulting tensors of the skip connections, that are used by predictions layers at different scales. Penalizing these tensors also indirectly affects the activity of encoder layers, as their output conditions the density of same-level encoder volumes because of the skip connections. Therefore, this regularization is less aggressive than penalizing all intermediary tensors and performances are less affected.

\begin{figure}[H]
    \centering
    \includegraphics[width=0.5\textwidth]{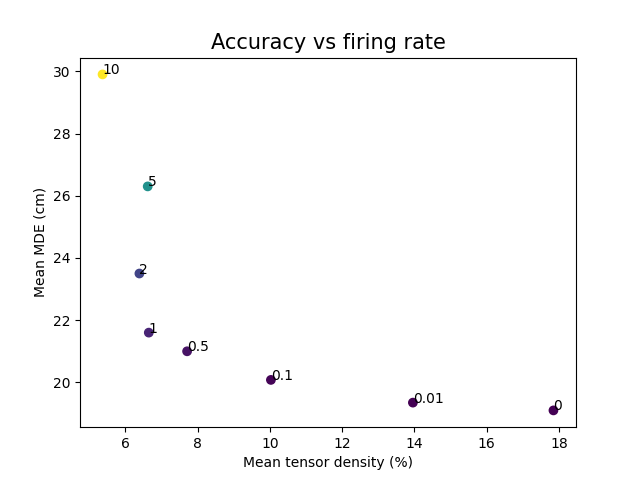}
    \caption{Test accuracy as a function of the mean firing rate. Mean firing rate is calculated as the average of the density of all activation (spike) tensors calculated within the network during inference. Labels correspond to the weight of the spike penalization loss in comparison to the objective loss. Unconstrained models generally perform better than models trained while encouraging sparsity. High weight values for the spike penalization loss do not always result, on average, in higher sparsity. Gradient descent on a non-convex problem does not always find the global minimum, and as such a penalization weight of 2 can result in both higher error and density than with a weight of 1.
    }
    \label{fig:acc_vs_rate}
\end{figure}

We then evaluate the network trained with spike penalization on the test set and compare obtained firing rates with our unconstrained model. More specifically, we plot the average test accuracy as a function of the average firing rate in Figure~\ref{fig:acc_vs_rate}. Regularized models show a drastic decrease in spiking activity, at a very low cost on the task accuracy. For instance, a network trained with a penalization weight of 0.1 sees a drop in MDE of less than 1 cm compared to the baseline (no penalization), but requires about 5 times fewer spikes. Densities less than 5\% are generally considered sufficient to leverage efficient sparse matrix operations. With these results, we can imagine our model implemented efficiently on dedicated hardware.\\

\section{Discussion} \label{discussion}

In this last section, we discuss several aspects concerning the computational efficiency of our model and its portability on current neuromorphic hardware.

\subsection{Target Hardware}
StereoSpike has resolutely been developed in the philosophy of spiking neural networks. As a result, it is essentially implementable on dedicated neuromorphic hardware, such as Intel Loihi \cite{Loihi}, IBM TrueNorth \cite{Truenorth}. These chips can leverage the binarity and sparsity of spike tensors navigating through the network. In addition, we believe that our model being feedforward and requiring a reset on all of its neurons at each timestep is not a problem, because resetting membrane potentials is actually less costly than applying a leak. Therefore, statelessness can be seen as an advantage over recurrence in spiking models with similar performances.
However, we are aware that current neuromorphic chips are initially designed for the implementation of stateful units, and acknowledge that we do not leverage this feature. Consequently, we believe that it rather fits to dedicated hardware for stateless models with sparse quantized activations. We therefore consider that Brainchip's Akida chip \cite{AkidaBrainchip} is a good fit. As it imposes weights to take at most 8 bit, we quantized StereoSpike's weights using PyTorch natively available post-training static quantization. The process resulted in an even lighter model with 8 bit wide unsigned integer weights, for the price of a minor performance drop (i.e., MDE of 17.1 cm on indoorflying split 1). Presumably, quantization-aware training would do even better. This demonstrates the efficient deployability on such hardware. Finally, we would like to emphasize that our class of model with sparse binary activations and less constrained weights provides a good compromise between Spiking Neural Networks (SNNs) and Binary Neural Networks (BNNs).

\subsection{Handling integer (non-binary) spike counts} Because of the sum operations present in residual layers of our architecture and at skip connections, bottleneck and decoder tensors can contain integer (non-binary) numbers of spikes. We explain here why we do not consider it as a problem.
On most digital neuromorphic chips, spikes are represented by multi-bit messages containing destination and/or source addressing, and a few bits for a graded-value payloads; this is the case for Loihi, see \cite{Davies2020}. In our case, the spike counts are included in $[0, 3]$ and thus can be coded on 2 bits. For the chips that can only handle binary spikes, a spike count of N could be handled by N serial binary spike operations. Another possibility to make our model fit in a small-size neuromorphic chip, could be to cut the skip connections as indicated in section \ref{skip_co}, or to remove intermediate prediction layers. The latter are indeed necessary to reach near state-of-the-art performances, but are not absolutely required to show reasonable performances.

\subsection{An estimated lower number of FLOPs}

The ever-expanding size of deep neural networks combined to the global energetic pressure has recently pushed researchers to use the number of Floating Point Operations (FLOPs) as a metric for algorithmic efficiency. Although this metric has been growing in popularity and would perfectly fit in StereoSpike's philosophy of sober AI, we decided not to include it in  Table \ref{table:deploy}.

The first reason is the lack of methodology regarding its estimation; the literature gives it a lot of attention, yet no conventions seems to have emerged from the community. Some libraries compatible with popular deep learning frameworks do exist \cite{flops_counter}, but we found out their estimations were highly flawed and only work for the most basic layer types.
A second reason is that because of the low bit-width of our activations, it would be unfair to count operations within StereoSpike as full 32-bit FLOPs. Furthermore, the Heaviside function that we use as activation throughout our network is intuitively much cheaper to compute than the more popular Sigmoid, Leaky ReLU, or GeLU activations; it is essentially a simple comparison with zero of a floating point number representing weighted spikes.

Yet, with all these concerns raised, and applying a worst-case estimation of the FLOPs necessary for one forward-pass of our model, we estimate it to require one order of magnitude less FLOPs than the average of our competitors. This feature is explained by the exclusive use of separable 2D convolutions, whereas other approaches have been built upon a backbone using full 3D convolutions.
As a result, we argue that StereoSpike is more computationally lightweight than the state-of-the-art.

\section{Conclusion and Future Work}

We have proposed StereoSpike, the first fully spiking, deep neural network architecture for large scale regression task with sparse activity. Lack of expressivity at the output has indeed hindered the development of SNNs on regression task, and we tackle this problem by increasing the number of spikes generated by deeper layers, with a pool of perfect integrator neurons bearing the final prediction with their membrane potential. The same strategy could presumably be used for other dense regression problems with SNNs, for example optic flow prediction.

The efficiency of our feedforward architecture combined with simple encoding has shown that depth estimation from DVS data can be brought back to a stateless, static, non-temporal task. As a result, hardware implementations of StereoSpike could consume substantially less than recurrent versions. However, we are aware that our model neither takes advantage of this implicit recurrence of SNNs to capture temporal dependencies, nor of the very high temporal resolution of DVSs; these two aspects deserve investigation for further improvement of our algorithm.

Furthermore, our experiments hint towards the fact that because of the constraint on their output (i.e. binarity) SNNs might generalize better than their ANN counterparts. Consequently, our work is yet another evidence that binary encoding in latent space can represent input data as efficiently as real-valued projections. This encourages more research in the field of SNNs and neuromorphic computing in general, or Binarized Neural Networks (BNNs).

Finally, the combination of event cameras and spiking neural networks within the same framework is a more biologically plausible approximation of the visual nervous system, and could allow researchers to understand the processing of depth in the brain with large-scale models.

%%%%%%%%% REFERENCES
{\small
\bibliographystyle{ieee_fullname.bst}
\bibliography{stereospike.bib}
}

\end{document}